\documentclass[runningheads]{llncs}
\usepackage{graphicx}
\usepackage[output-decimal-marker={.},separate-uncertainty=true]{siunitx}
\usepackage{subcaption}
\usepackage{wrapfig}
\usepackage{hyperref}

\usepackage[colorinlistoftodos]{todonotes}
\usepackage[markup=bfit, deletedmarkup=sout, authormarkup=brackets]{changes}
\usepackage[xindy]{glossaries} 
\definechangesauthor[name={Hagen}, color=blue]{HL}
\definechangesauthor[name={Adam}, color=green]{AR}
\definechangesauthor[name={Matej}, color=orange]{MH}

\usepackage{draftwatermark}

\begin{document}
\SetWatermarkAngle{0}
\SetWatermarkColor{black}
\SetWatermarkLightness{0.5}
\SetWatermarkFontSize{10pt}
% \SetWatermarkScale
% \SetWatermarkHorCenter
\SetWatermarkVerCenter{25pt}
\SetWatermarkText{\parbox{30cm}{%
\centering This is the authors' final version of the manuscript published as\\
\centering Lehmann, H., Rojik, A., Friebe, K., Hoffmann, M. (2022). Hey, Robot! An Investigation of Getting Robot’s Attention Through Touch. \\
\centering In: Social Robotics. ICSR 2022. Lecture Notes in Computer Science, vol 13817. Springer, Cham. \\
\centering \url{https://doi.org/10.1007/978-3-031-24667-8_35}
}}

\title{Hey, robot! An investigation of getting robot's attention through touch\thanks{This work was supported by the Czech Science Foundation (GA {\v C}R), project no. 20-24186X. H.L. was supported by the International Mobility of Researchers in CTU, Nr. CZ.02.2.69\/0.0\/0.0\/16\_027\/0008465.}}

%\titlerunning{Getting robot's attention through touch}
\author{Hagen Lehmann\inst{1,2}\orcidID{0000-0001-9676-2269} \and Adam Rojik\inst{2} \and Kassandra Friebe\inst{3,4} \and
Matej Hoffmann\inst{2}\orcidID{0000-0001-8137-3412}}
\authorrunning{H. Lehmann et al.}
\institute{Università di Bergamo, Dipartimento di Scienze Umane e Sociali
\email{hagen.lehmann@unibg.it}\\
%\url{http://www.springer.com/gp/computer-science/lncs} 
\and
Department of Cybernetics, Faculty of Electrical Engineering \\ Czech Technical University in Prague, Czech Republic\\
\email{matej.hoffmann@fel.cvut.cz}
\and
Faculty of Mathematics, Physics and Informatics, Comenius University\\
Bratislava, Slovakia
\and 
Department of Cognitive Science, Central European University, Vienna, Austria
}
\maketitle              % typeset the header of the contribution
%
%The abstract should briefly summarize the contents of the paper in 15--250 words.
\begin{abstract}
Touch is a key part of interaction and communication between humans, but has still been little explored in human-robot interaction. In this work, participants were asked to approach and touch a humanoid robot on the hand (Nao -- 26 participants; Pepper -- 28 participants) to get its attention. We designed reaction behaviors for the robot that consisted in four different combinations of arm movements with the touched hand moving forward or back and the other hand moving forward or staying in place, with simultaneous leaning back, followed by looking at the participant. We studied which reaction of the robot people found the most appropriate and what was the reason for their choice. For both robots, the preferred reaction of the robot hand being touched was moving back. For the other hand, no movement at all was rated most natural for the Pepper, while it was movement forward for the Nao. A correlation between the anxiety subscale of the participants' personality traits and the passive to active/aggressive nature of the robot reactions was found. Most participants noticed the leaning back and rated it positively. Looking at the participant was commented on positively by some participants in unstructured comments. We also analyzed where and how participants spontaneously touched the robot on the hand. In summary, the touch reaction behaviors designed here are good candidates to be deployed more generally in social robots, possibly including incidental touch in crowded environments. The robot size constitutes one important factor shaping how the robot reaction is perceived. 
\end{abstract}

\section{Introduction}
People constantly use a variety of nonverbal cues while interacting. These include gaze and eye movement, gesture, mimicry and imitation, touch, posture and movement, interaction rhythm and timing \cite[Chapter 6]{bartneck2020}. For successful human-robot interaction (HRI), machines should be able to understand as well as produce these cues. According to \cite{breazeal2016social}, ``humans are born as \textit{tactile creatures}. Physical touch is one of the most basic forms of human communication.'' Even a single gentle touch can have important effects as demonstrated by the so-called ``Midas touch'' effect \cite{crusco1984}, for example. At the same time, physical contact is a very intimate type of interaction that is to be used with caution. Heslin~\cite{heslin1983} studied the pleasantness vs. intrusiveness of touch between sexes in relation to a stranger, a friend, and a close friend in the United States. The body areas where touch was rated as pleasant or unpleasant strongly depend on the combination of these factors. 
In HRI, touch has still been relatively unexplored to date.

The bottleneck of wider use of touch in HRI has largely been technological---robust and affordable tactile sensors were lacking. Recently, the number of available technologies is growing (see e.g., the 2019 special issue of Proceedings of the IEEE \cite{Dahiya2019}). The focus has largely been on tactile sensing for manipulation, as equipping only robot hands/fingers with tactile sensors requires relatively small patches of electronic skin. Whole-body artificial skins for robots have been an exception, with only a few successful technologies deployed on complete robots and over extended time periods (e.g., \cite{maiolino2013} on the iCub humanoid or the multimodal skin of Mittendorfer, Cheng, and colleagues \cite{Mittendorfer2011}). 
Large patches of sensitive skin are also important for the safety of robot manipulators, as demonstrated by Airskin, for example (see \cite{Svarny2022Airskin}). Silvera-Tawil et al.~\cite{silveratawil2015} provide a review of artificial skin and tactile sensing for socially interactive robots. For example, tactile sensors on the therapeutic seal robot PARO are an important sensory modality supporting its interactions \cite{shibata2012therapeutic}. The Nao and Pepper social humanoids use capacitive touch sensors on the head and arms to detect human contact.

There are several ways in which touch enters HRI (see \cite{argall2010,shiomi2020} for surveys). The first division comes from who initiates the contact. More studies investigated the case where the robot initiated contact \cite{arnold2018,chen2011touched,chen2014investigation,van2013touch,willemse2016observing,zheng2020}. Humans touching robots has been less explored \cite{alenljung2018,robins2010developing,robins2012embodiment}. Cramer et al.~\cite{cramer2009} had participants rate videos of both interaction types and found that communicative touch could be considered a more appropriate behavior for proactive agents rather than reactive agents. Affective touch, as manifested by the PARO robot \cite{shibata2012therapeutic}, constitutes a different context compared to communicative touch. Finally, while most studies focus on deliberate touch, incidental contacts like in a crowded environment and reactions to them \cite{garcia2019} are also relevant for HRI.

In this work, participants were instructed to get the attention of a humanoid robot (Nao or Pepper) looking to the side by touching its hand. The hand was chosen as touch on this body part is the most acceptable across different contexts (gender, stranger / friend) \cite{heslin1983}. The robot reactions were preprogrammed and consisted in four different combinations of arm movements with the touched hand moving forward or back and the other hand moving forward or staying in place, accompanied by leaning back and followed by looking at the participant. The goal was to design behaviors that would be perceived as natural by participants. We were loosely inspired by prank videos like the ``Touching Hands on Escalator Prank'', \href{https://youtu.be/BTl5HC9VfAE}{https://youtu.be/BTl5HC9VfAE}. We studied: (i) which reaction of the robot participants found the most appropriate and what was the reason for their choice;  (ii) where and how participants touched the robot on the hand. With the focus on designing and assessing robot reactions to touch, this study provides a new contribution to the field, complementing the work of Shiomi et al.~\cite{shiomi2018} who investigated pre-touch reactions.
\section{Related Work}
Here we review previous studies specifically relevant to our scenario, structured as follows: robots touching humans, humans touching robots, and robot reactions to touch. 

\subsection{Robots touching humans}
Guidelines for robots how to touch humans are provided in \cite{van2013touch}.
Chen et al.~\cite{chen2011touched,chen2014investigation} studied responses to robot-initiated touch in a nursing context and found that instrumental touch---cleaning the person's skin---was perceived more favorably than affective touch (providing comfort). Zheng et al.~\cite{zheng2020} used the female android Erica to touch participants' hand or finger, varying the touch type (contact or pat) and the duration and intensity of the contact. They studied the effect on participants' arousal and whether the robot succeeded to communicate a specific emotion through the touch.  

Other studies used videos, with participants rating sequences in which humans were touched by a robot. Arnold and Scheutz~\cite{arnold2018} using the PR2 robot found that touch improves people's evaluation of a robot's
social performance. In \cite{willemse2016observing}, participants rated the pleasantness of touch when a person on the video was touched on the hand by a human hand, Nao robot hand, mannequin arm, or plastic tube. Stroking touches with a velocity of ca. 3 cm/s were rated as most pleasant.
Robot touch was not rated as significantly more pleasant than either touches applied with the mannequin hand or tube.

\subsection{Humans touching robots}
The studies in which humans touch robots are of two main types. The first group is constituted by studies in which participants are asked to touch a robot in a specific fashion. In \cite{alenljung2018}, people were asked to  touch a Nao robot, expressing one of eight emotions: anger, disgust, fear, happiness, sadness, gratitude, sympathy, or love.
In \cite{burns2022endowing}, specific instructions about affective touch
communication gestures (hitting, poking, squeezing, stroking, and tickling) at two force intensities (gentle and energetic) and sensor locations were provided to participants.

In a second group of studies, humans touched robots spontaneously. Robins et al.~\cite{robins2010developing,robins2012embodiment} studied interaction of autistic children with the robot KASPAR and identified grasping, stroking and poking. The intensity of touch varied between tight or firm and very light or `gentle'. In \cite{garcia2019}, contacts were not deliberate but incidental, as a Pepper robot was moving trough a crowd. Impact, push, and clamp were the touch types identified and 70\% of them were with the arms and hands of the robot. 

\subsection{How should the robot respond?}
In many studies on social touch in HRI, experiments end their evaluation with the moment of touch and the reaction to touch and its appropriateness is not assessed (e.g., \cite{alenljung2018,burns2022endowing}). Shiomi et al.~\cite{shiomi2018} studied pre-touch reactions. As participants were approaching the face of a female android with their hand, the robot displayed a reaction---turning the head toward the person---either at 45 or 20 cm before contact (pre-touch) or only after contact. The pre-touch reaction at 20 cm was rated most positively. For the KASPAR robot, several responses were designed: for example, touching the robot hand caused the robot to raise its hand; touching the shoulder caused the robot to move the arm to the side \cite{robins2010developing}. The robot responded to `aggressive' tactile interaction by displaying its `sad' expression, face covered by hands, evasive body movements, or by saying ``ouch – this hurts''  \cite{robins2012embodiment}. Garcia et al.~\cite{garcia2019} experimented with a Pepper robot in a crowd. The impacts (contact, push, clamp) were not detected by tactile sensors but through joint torques. The reactions consisted in local or whole-body compliant behaviors in the direction of the impact.  

Different than in the works described above, we designed the following responses to touch on the robot's hand: (1) look at the touched hand; (2) lean back; (3) move the arms in 4 different ways. Rather than asking the participants about their perception of the robot (e.g., \cite{arnold2018}) or pleasantness or appropriateness of the situation involving touch in general (e.g., \cite{chen2011touched,cramer2009,willemse2016observing,zheng2020}), we specifically asked them to rate the robot reactions.

\section{Methods}
\label{sec:methods}

\subsection{Participants}
There were 26 participants (13 female; mean age 29.8 years; ranging from 21 to 68) interacting with the Nao robot and 28 participants (12 female; mean age 28.5, ranging from 20 to 53) interacting with the Pepper robot. 
On a 5-point Likert scale concerning their experience with robots and ranging from 1 = no experience to 5 = very experienced, the ``Nao group'' reported an average experience with robots of 2.3; the ``Pepper group'' 2.1. 
Participants were recruited from Facebook local area groups, experimenters' social circles such as family and friends, and from the administrative personnel of the Department. 

\subsection{Nao and Pepper humanoid robots}
Both robots were manufactured by Aldebaran, now Softbank Robotics. They use the same middleware (NAOqi) and programming environment. The robot movements used in this work were designed in Choregraphe. 

\subsubsection{Nao robot.} 
\begin{figure}[!ht]
    \centering
    \begin{subfigure}[b]{0.2\textwidth}
        \includegraphics[width=\textwidth]{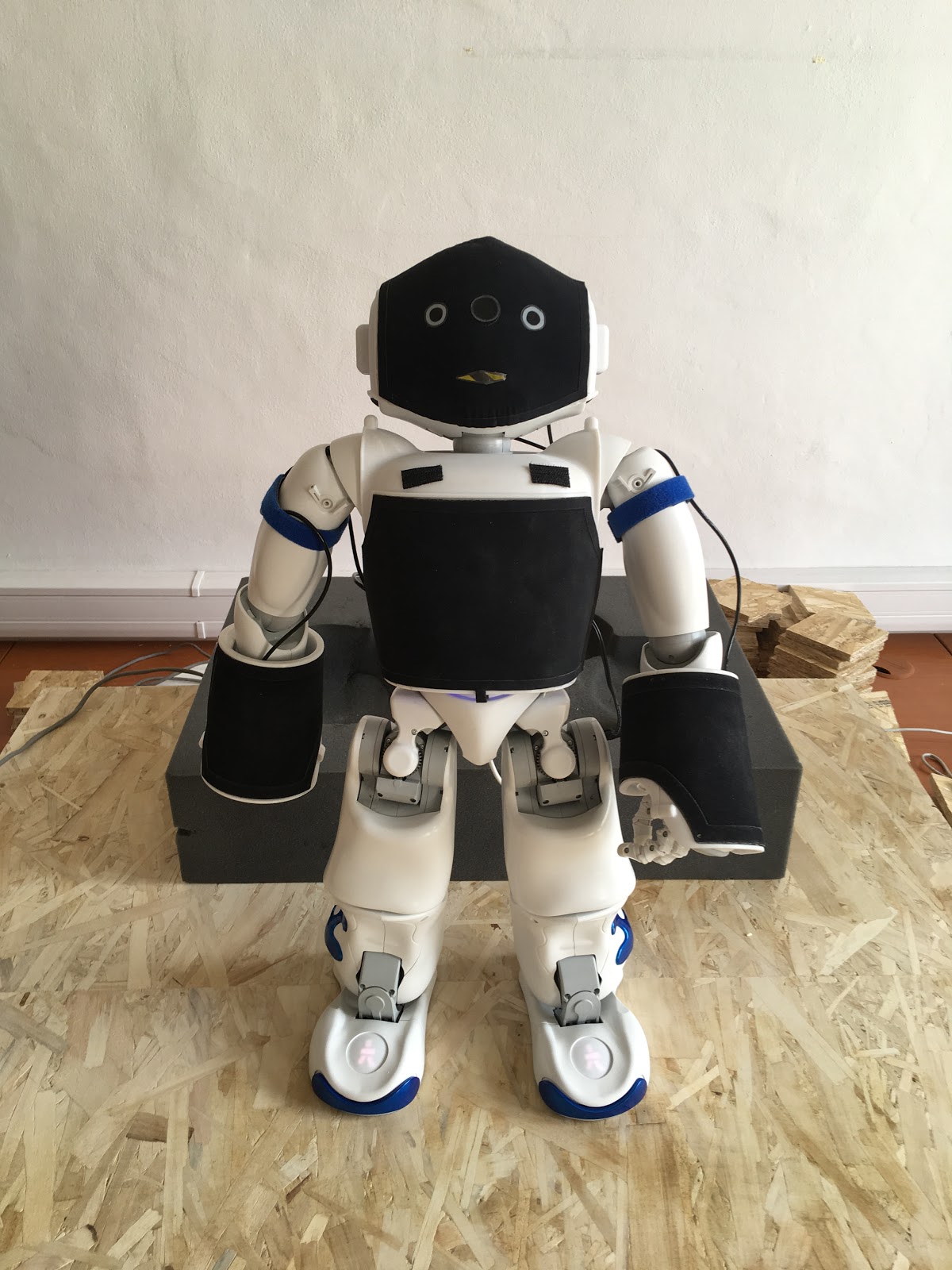}
        \caption{}
    \end{subfigure}
    \hfill
    \begin{subfigure}[b]{0.2\textwidth}
        \includegraphics[width=\textwidth]{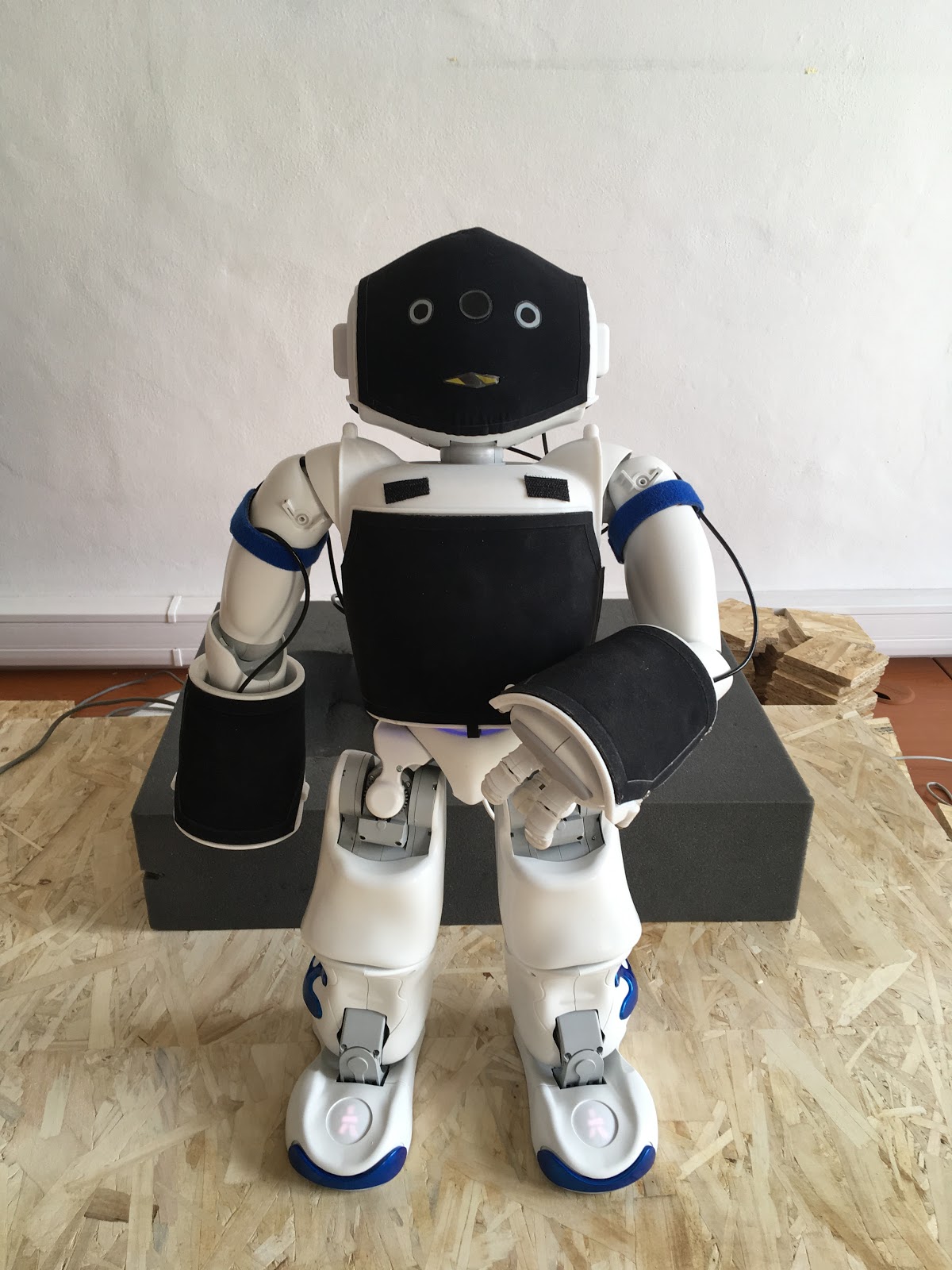}
        \caption{}
    \end{subfigure}
    \hfill
    \begin{subfigure}[b]{0.2\textwidth}
        \includegraphics[width=\textwidth]{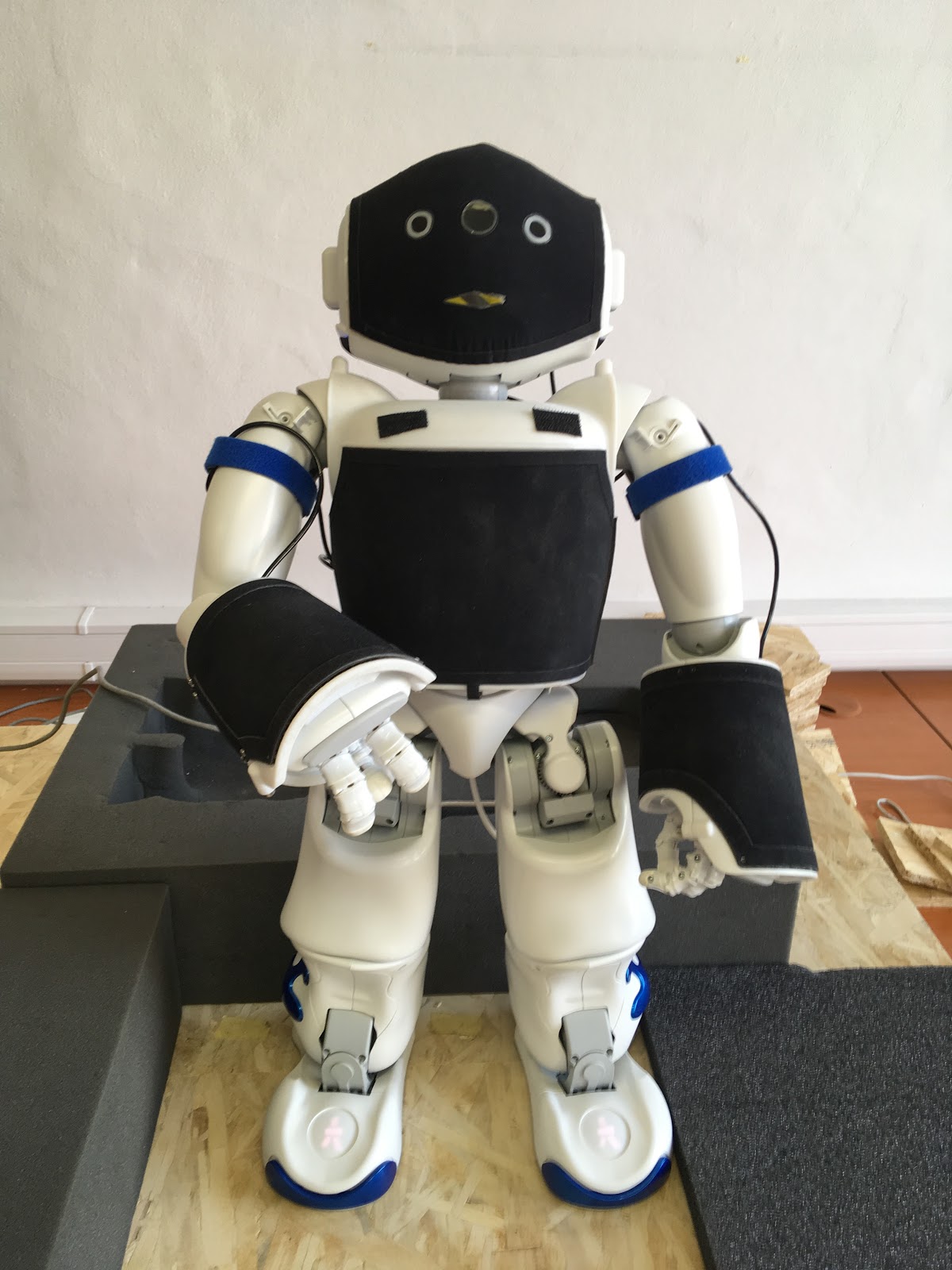}
        \caption{}
    \end{subfigure}
    \hfill
    \begin{subfigure}[b]{0.2\textwidth}
        \includegraphics[width=\textwidth]{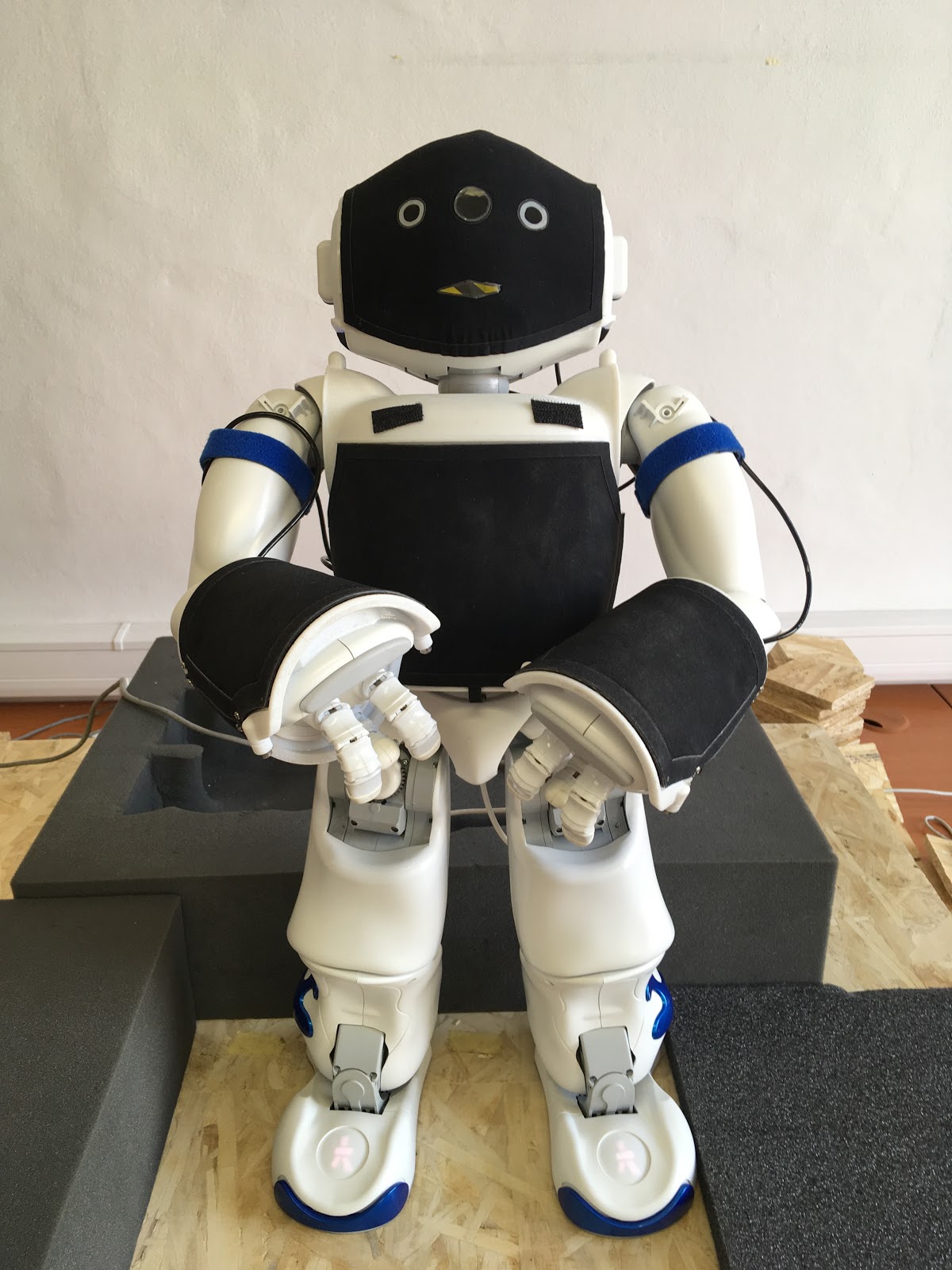}
        \caption{}
    \end{subfigure}
    \caption{Nao -- arm movements for startle reaction. (a) Touched hand back, other hand still. (b) Touched hand back, other hand forward. (c) Touched hand forward, other hand still. (d) Touched hand forward, other hand forward. Other components of the behavior (gaze, lean back) not shown here. }
    \label{fig:nao-startle}
\end{figure}

We used Nao version Evolution V5 (H25 V50). Our robot exemplar was uniquely equipped with artificial sensitive skin (black parts on Fig.~\ref{fig:nao-startle}), making the robot slightly taller ($\SI{59}{\cm}$ compared to $\SI{57.4}{\cm}$) than the original. The skin is a capacitive tactile system commonly used on the iCub robot~\cite{maiolino2013} and custom-designed for the Nao robot.  

\subsubsection{Pepper robot.} 
\begin{figure}[!htb]
    \centering
    \begin{subfigure}[b]{0.2\textwidth}
        \includegraphics[width=\textwidth]{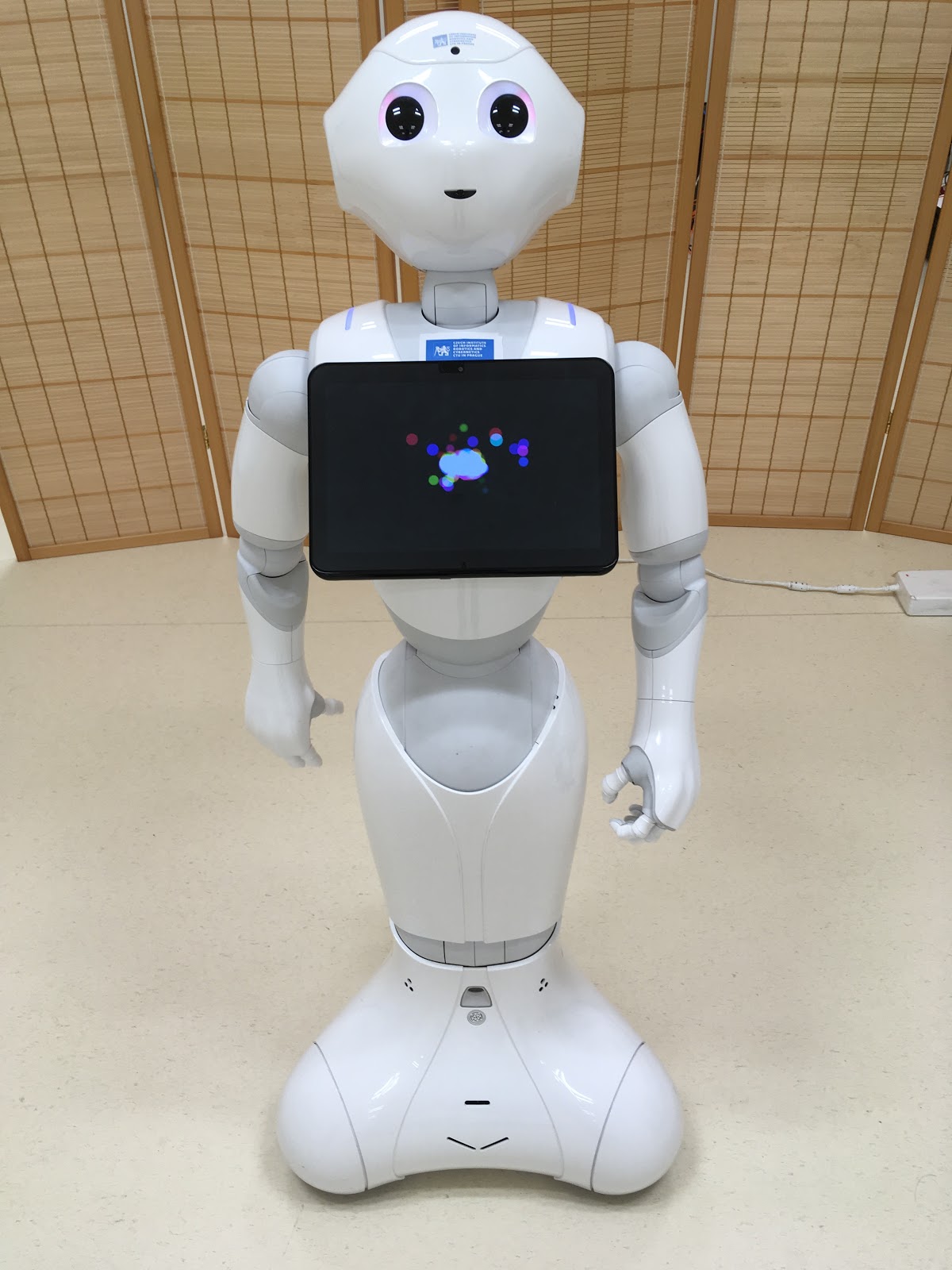}
        \caption{}
    \end{subfigure}
    \hfill
    \begin{subfigure}[b]{0.2\textwidth}
        \includegraphics[width=\textwidth]{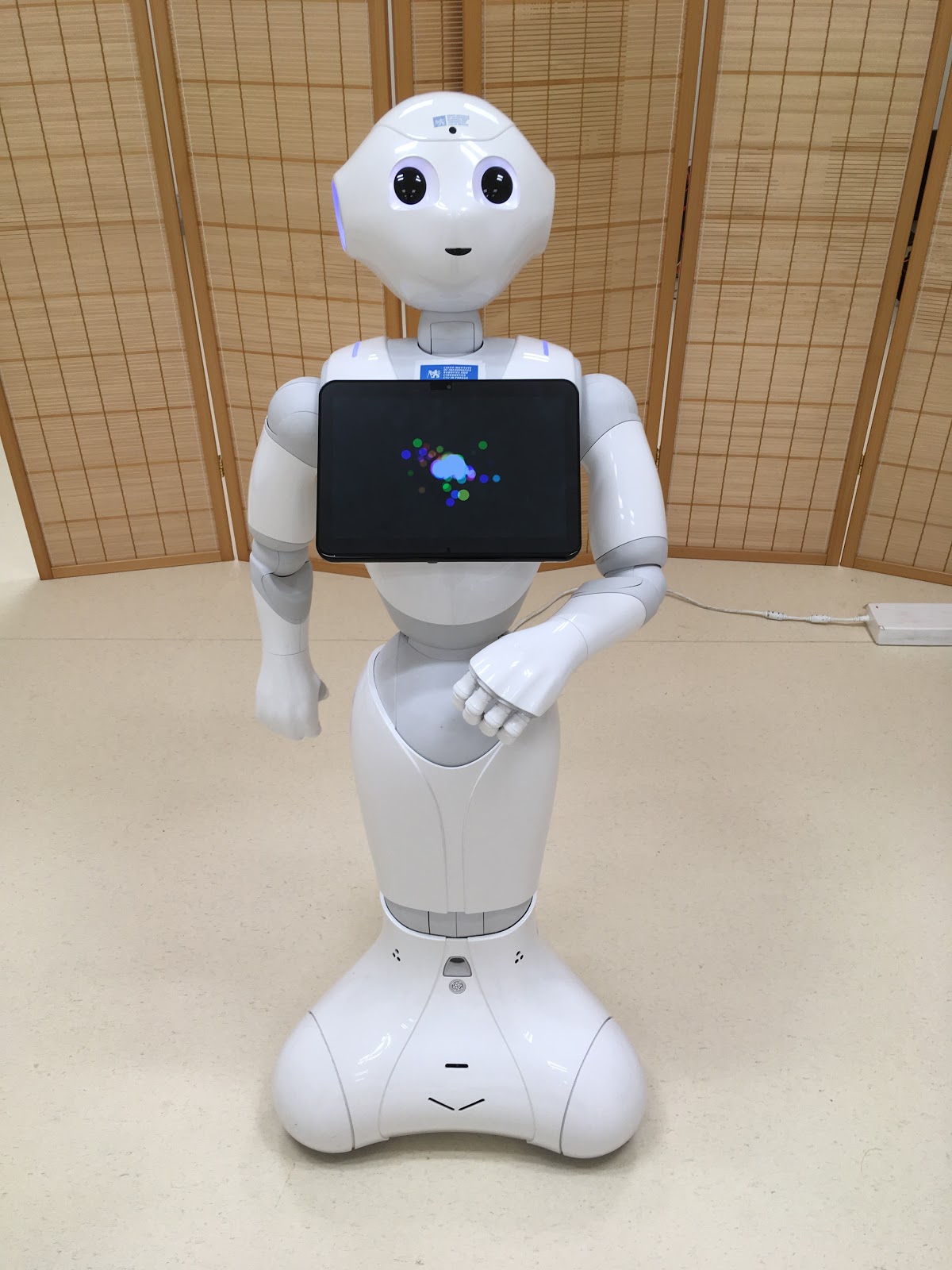}
        \caption{}
    \end{subfigure}
    \hfill
    \begin{subfigure}[b]{0.2\textwidth}
        \includegraphics[width=\textwidth]{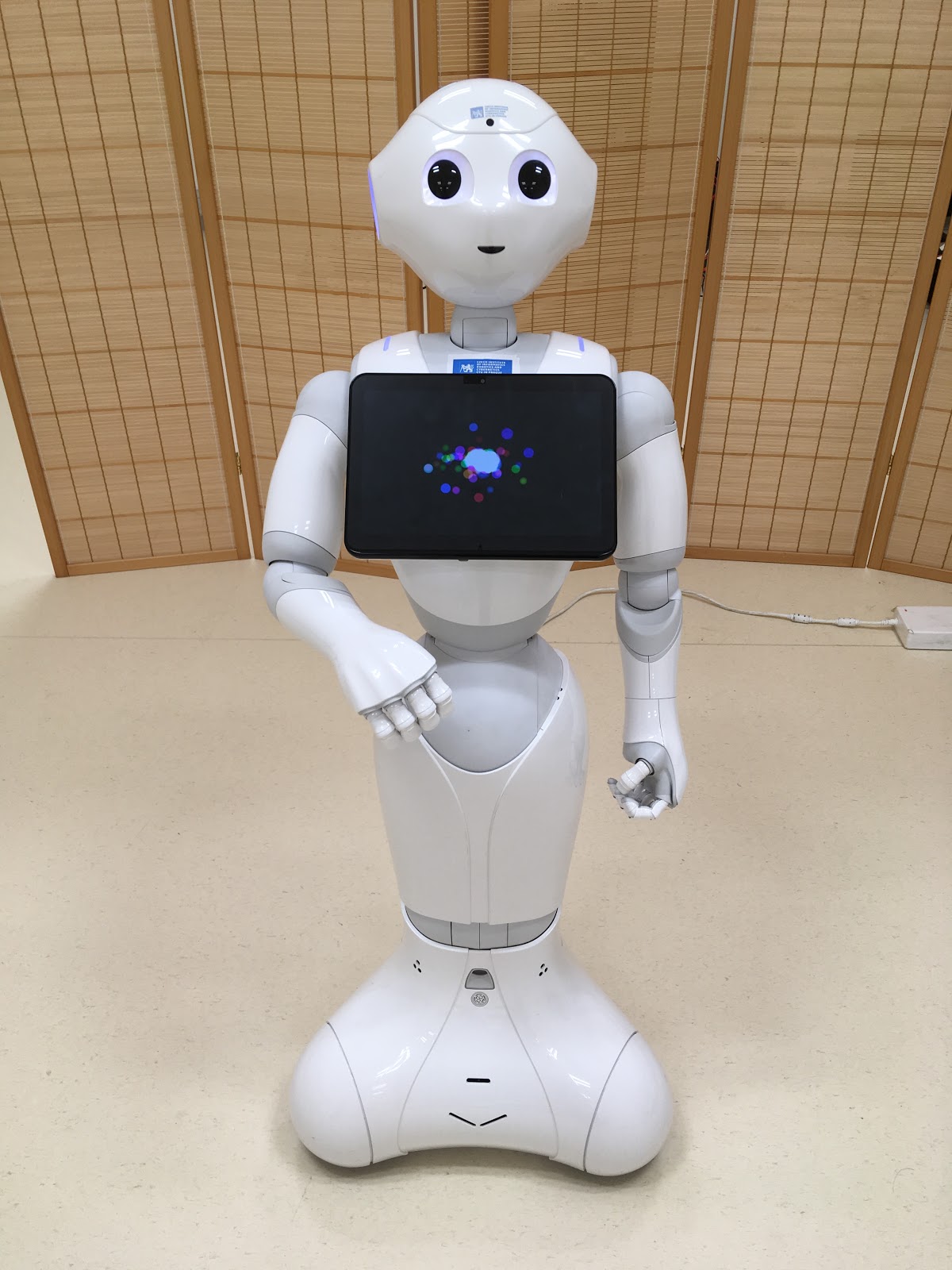}
        \caption{}
    \end{subfigure}
    \hfill
    \begin{subfigure}[b]{0.2\textwidth}
        \includegraphics[width=\textwidth]{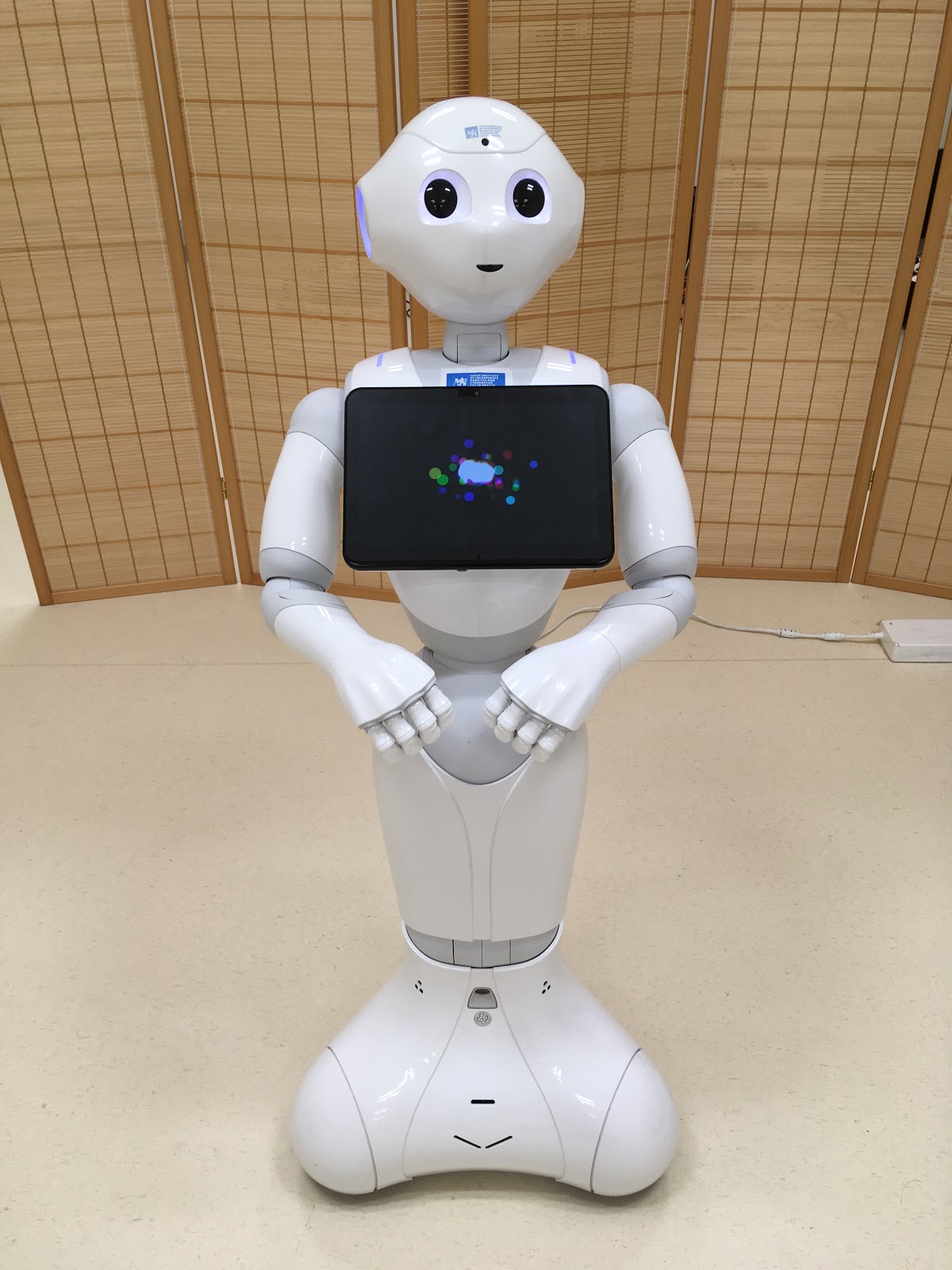}
        \caption{}
    \end{subfigure}
    \caption{Pepper -- arm movements for startle reaction.  (a) Touched hand back, other hand still. (b) Touched hand back, other hand forward. (c) Touched hand forward, other hand still. (d) Touched hand forward, other hand forward.  Other components of the behavior (gaze, lean back) not shown here.}
     \label{fig:pepper-startle}
\end{figure}

Humanoid robot Pepper (version 1.8a) was used (Fig.~\ref{fig:pepper-startle}). It is $\SI{120}{\cm}$ tall and has touch sensors on its arms.

\subsection{Touch detection}
Touch on the robot arm is detected differently for the two robots. On the Pepper, the touch sensors on the outer part of its hand were used. On the Nao, the capacitive pressure-sensitive skin surrounding the whole hand and wrist and comprising 240 sensors was used (Fig.~\ref{fig:nao-startle}). To detect and then respond to touch in real time, the asynchronous calls available in NAOqi API 2.4.3 were used.\footnote{NAOqi API 2.4.3 was standard in the Pepper robot. In the Nao robot used (NAOqi API version 2.1.4) the newer NAOqi API version had to be used in addition to allow for these calls to be used.}  

In the Pepper robot, touch sensors' output was true or false and could be directly registered to trigger a response. In our Nao, sensitive skin outputs were sensed by the YARP middleware. Contact detection was then performed by a module running in a separate thread, comparing the signals with a threshold value---if crossed, a contact event was triggered.  

\subsection{Reactions to touch}
\label{subsec:reactions_to_touch}
A key contribution of this work is the design of robot reactions to unexpected touch. We took inspiration from how humans react in such situations (e.g., \href{https://youtu.be/BTl5HC9VfAE}{https://youtu.be/BTl5HC9VfAE}) and we tested several behaviors on ourselves. An illustration of the behaviors we developed is available in the accompanying video \href{https://youtu.be/TI_oy6uO0Kw}{https://youtu.be/TI\_oy6uO0Kw}.

\subsubsection{Arm movements.} 
\begin{figure}[!htb]
    \centering
        \includegraphics[width=\textwidth]{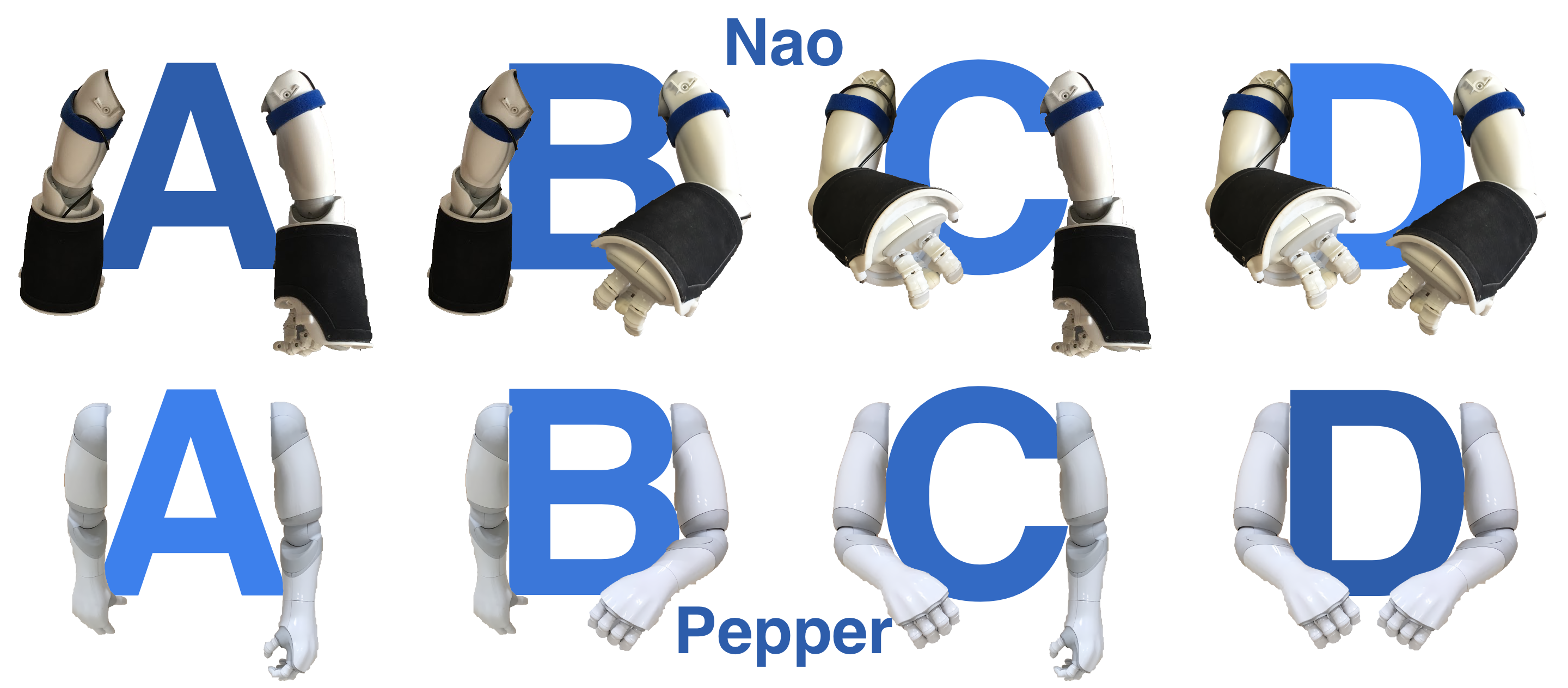}
        \caption{Startle reactions -- arm movements schematics. (Top) Nao, (Bottom) Pepper. (A) Touched hand back, other hand still. (B) Touched hand back, other hand forward. (C) Touched hand forward, other hand still. (D) Touched hand forward, other hand forward. See Figs \ref{fig:nao-startle},\ref{fig:pepper-startle} for complete robot photos.}
        \label{fig:reactions-ABCD}
\end{figure}

Since it was the robot hand that was unexpectedly touched, some reaction with the contacted limb seems natural. An instinctive reaction appears to be to retract the arm / hand that was unexpectedly contacted (Fig.~\ref{fig:reactions-ABCD}, A). This may be accompanied by moving the contralateral arm forward (B), constituting a defensive reaction. However, people may not find a robot defensive reaction the most appropriate and could perceive it as detached. We thus added also the other two possible combinations of arm movements with the touched hand moving forward and the other hand remaining still (C) or moving forward as well (D). Those movements were created in Choregraphe and then exported into Python code.
We experimented with additional behaviors like raising the touched hand or both hands to shoulder height, but these movements appeared too aggressive and were omitted.

\subsubsection{Lean back.}
Movements of the whole body are more natural than isolated arm movements. After experimentation, we accompanied the arm movements with simultaneous leaning back of the torso, taking advantage of the implementation in \cite{Lehmann_Proxemics_2020}. 

\subsubsection{Gaze.}
Furthermore, when confronted with an unexpected sensory percept such as a touch, it is natural to seek the cause of the stimulation. One possibility would be to look at the touched hand. After initial experimentation, we decided to design this component of the reaction differently. Rather than adding to the immediate startle reaction (move arms, lean back), we decided to add a slower head movement that directs the gaze on the participant, looking approximately into his eyes. This seems a more natural reaction in case the unexpected touch was not threatening. In the Pepper robot, this was accompanied by blinking. 

\subsection{Experimental procedure}
\label{subsec:questionnaires}
Before each experiment, the participants filled in an informed consent form.
With the goal of assessing the personality dimensions of our participants we asked them at the beginning of the experiment to complete the Ten Item Personality Inventory (TIPI) questionnaire, a brief measure of the ``Big-Five'' personality dimensions \cite{gosling2003very}.\footnote{Prof. Marek Fran{\v e}k kindly provided the Czech version \cite{sefara2015socio}.}  The participants were then instructed to approach the robot (Nao or Pepper) and touch the robot’s right hand. The robot was initially looking to the other side and did not know about the participant. After the robot detected the touch, it displayed one of the four reactions (Section~\ref{subsec:reactions_to_touch}). 
The scenario was repeated four times, with the order of startle reactions randomized for individual participants. After this, the participants filled in a custom questionnaire choosing which reaction they liked best, followed by questions whether they noticed and liked the leaning back. Finally, participants had the opportunity to add any comments regarding their general impression from the experiment.

\section{Results}
\label{sec:results}

\subsection{Which reaction was the most appropriate?}
The main result consists in the assessment of the four reaction types by the participants. A summary is in Table~\ref{tab:experiments-result} below. 
For the Nao robot, the preferred reaction type was B---the touched hand moving back and the other hand moving forward. For the Pepper, the preferred reaction was the touched hand moving back and the other hand remaining still.
Taking both robots together, the reactions involving the touched hand moving backward (A, B) were preferred over those where the touched hand moved forward (C, D). 
We performed a $\chi^2$ test comparing the number of A and B together versus C and D with the expected probability distribution 0.5 for each. The result did not yield statistical significance ($\chi^2 = 2.3$, $df=1$, $p=0.13$).   

\begin{table}[]
\centering
\begin{tabular}{lrrrr}
\hline
\multicolumn{1}{|l|}{\textbf{Robot \textbackslash Reaction}} &
  \multicolumn{1}{l}{\textbf{A}} &
  \multicolumn{1}{l}{\textbf{B}} &
  \multicolumn{1}{l}{\textbf{C}} &
  \multicolumn{1}{l|}{\textbf{D}} \\ \hline
\textbf{Nao}    & 6           & 10          & 6           & 4           \\
\textbf{Pepper} & 11          & 6           & 5           & 6           \\
\textbf{both robots} & \textbf{16} & \textbf{16} & \textbf{11} & \textbf{10}
\end{tabular}
\caption{Most appropriate reaction reported by participants. For example, the top left number means that 6 participants interacting with the Nao liked the reaction ``A'' best.}
\label{tab:experiments-result}
\end{table}

Further, we evaluated the free text answers from the questionnaires where the participants commented on their choices. Most participants advocated their choice that the chosen reaction was the most natural, appropriate, or logical to them.  Some mentioned that this is what ``they would do'' in such a context. Some participants reported that the most natural reaction depends on the context. If unexpectedly touched by a stranger (our case), a more defensive reaction (A,B) is expected. If the robot was touched by a friend and not taken by surprise, reaction involving movement toward the participant (C or D) would be natural. There were also some qualitative differences between the words the two participant groups used, which could be attributed to the different robot size. For the Nao robot, some participants that opted for C or D reported that they found the reactions A or B too fearful---the robot appears frightened and unfriendly and does not want to interact with them. Reactions C and D were found more interactive and friendly by some. For the Pepper group, the word ``aggressive'' or its synonyms was more present in the dictionary, mostly in relation to assessing reactions C or D that involved forward movements toward the participant. 

Most participants reported in the questionnaire that they noticed the leaning back component of the behavior (21/26 for the Nao; 26/28 for the Pepper) and most evaluated it positively (18/26 for Nao; 22/28 for Pepper).

\subsection{Interaction with personality traits and experience with robots}

\subsubsection{Correlation of Anxiety scores with preferred reaction type.}
The robot reactions (Fig.~\ref{fig:reactions-ABCD} for an overview of the arm movements) can be approximately ordered from the most passive (A -- touched hand moves back), over less passive (B -- touched hand moves back, other hand moves forward) to more active or perhaps even ``aggressive''---the touched hand moves forward (C) and both hands move forward (D). We hypothesised that the assessment of the reactions by the participants could correlate with their anxiety score as reported by the TIPI subscale 'Anxious, easily upset.' We hypothesized that participants with higher anxiety scores may prefer more passive robot reactions. Spearman's rank correlations were computed to assess this relationship. A negative correlation between participants' anxiety and the ``aggressiveness'' of the robot reaction, i.e. people with higher anxiety scores avoiding aggressive and preferring more passive robot reactions was found for the Pepper robot (r(26) = -0.25, p-value = .202, not significant). Interestingly, for the Nao robot, a significant positive correlation (r(24) = 0.49, p = .011) was found. One can speculate that the robot size makes the difference here. For the Nao robot that is small and perceived as not threatening, participants with higher anxiety scores may not be concerned about their safety, but instead may fear that the robot will not interact with them and hence they may prefer more active robot reactions. 

\subsubsection{Correlation of Experience with preferred reaction type.}
A Spearman's rank correlation did not reveal any significant correlation of participants' experience with robots and their preferred reaction, r(52) = -0.09, p-value = .518.

\subsection{Touch location and type}
\begin{figure}[!htbp]
    \centering
    \begin{subfigure}[b]{\textwidth}
       \includegraphics[width=\textwidth]{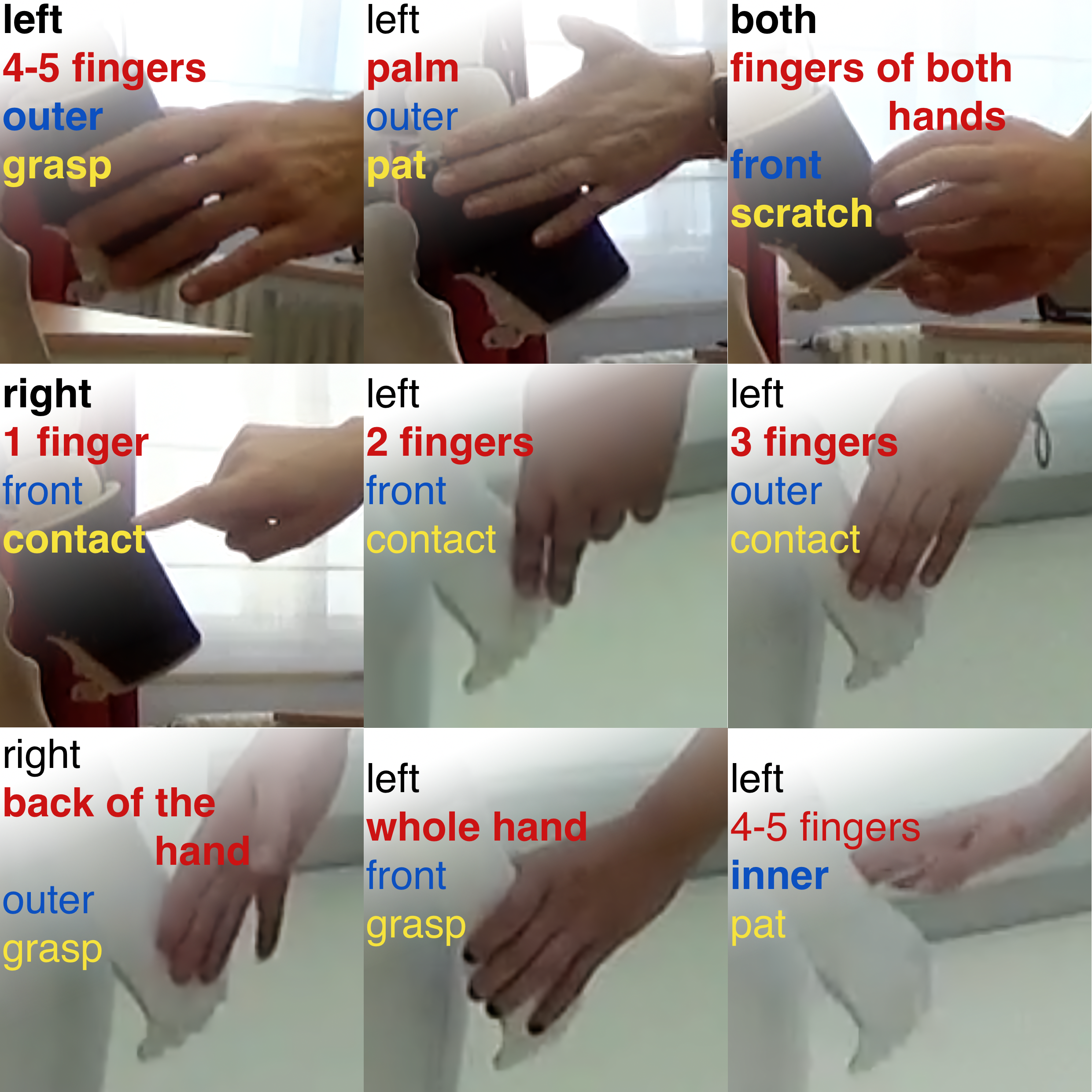}
        \caption{}
        \label{fig:touch-styles-all}
    \end{subfigure}
    \begin{subfigure}[b]{\textwidth}
        \includegraphics[width=0.45\textwidth]{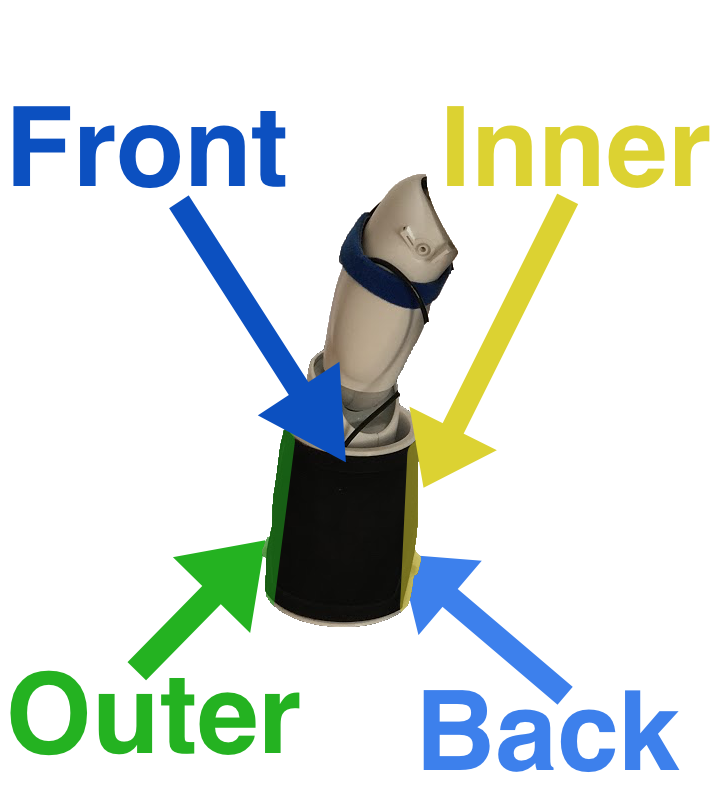}
        \includegraphics[width=0.45\textwidth]{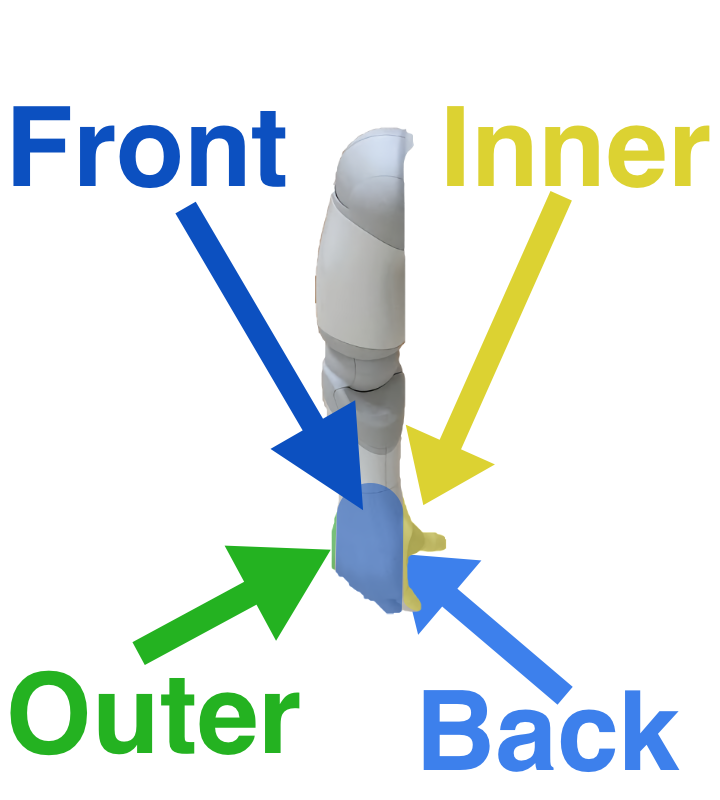}
        \caption{}      
        \label{fig:touch-on-robot-hand-location}
    \end{subfigure}
    \caption{a) Touch location and type -- examples. We distinguished which hand the participants used for contact (left/right/both), what part of their hand (1,2,3,4-5 fingers, palm, whole hand, or back of the hand), what part of the robot hand was contacted (front/back/inner/outer), and the type of contact (contact/grasp/pat/scratch). b) Touch location on robot hand (Left) Nao (Right) Pepper.}
 \end{figure}

How touch is delivered---location, duration, intensity, type---importantly modifies this act of communication and allows to express different emotions, for example (see \cite{hertenstein2009communication} and \cite{alenljung2018} for application to HRI). In this work, the participants were only instructed to get the robot's attention by touching its right hand. After contact was detected, the robot response was triggered so the touch duration could not be studied. Intensity was not available from the sensors on the Pepper; the skin our Nao is retrofitted with could detect pressure values, but in our experiments, it was binarized. Therefore, we used the video recordings of the experiments to study: (i) which hand participants used to contact the robot hand; (ii) which part of their hand they used for contact; (iii) which part of the robot hand did they contact; (iv) what was the type of touch (e.g., brief contact, grasp). Due to a technical problem (full memory card), some videos with the Nao robot were not saved. Thus, in this section, 15 Nao participants and all 28 Pepper participants' interactions are analyzed. 
Examples of the different combinations are in Fig.~\ref{fig:touch-styles-all}.

\subsubsection{Participants' hand and location.}
As the participants were instructed to approach the robot that was facing them and touch the robot's right hand in a habitual resting posture next to the robot's body, it was natural for them to use their left hand to touch the robot. Most people indeed used their left hand (80\% with Nao; 85.7\% with Pepper). The rest used the right hand; one participant used both hands to contact the Nao robot.

Regarding the part of the participants' hand used to touch the robot, more than 60\% of the participants used their fingers (1, 2, 3, or 4-5) to contact the robot. The rest used the palm or the whole hand. 

\subsubsection{Location on robot's hand.}
Given the nature of the Nao's hand in our case---basically a cylinder---we limited the classification to whether the participants touched the front, back, inner, or outer part of the robot hand. The same was applied to the Pepper. See Fig.~\ref{fig:touch-styles-all}. As would be expected, the vast majority of the participants touched the outer or frontal part of the robots' hands.

\subsubsection{Contact type.}
Finally, we classified the type of contact---how participants touched the robot. The majority of the participants used a brief, ``atomic'', touch or contact (62.5\% for Nao; 75\% for Pepper). Some participants have grasped the robot hand (12.5\% Nao; 14.3\% Pepper). Pat or scratch was also used occasionally.

\section{Conclusion, Discussion, Future Work}

In this study, we presented participants with a new type of HRI scenario involving touch, in which they were asked to approach a robot (Nao or Pepper) looking elsewhere and get its attention by touching its hand. The main contribution of this work is the focus on the design and participants' assessment of the robot reactions. The key component of the robot response were four different combinations of arm movements. For both robots, most participants preferred when the touched robot hand moved backwards in response to contact. Such reactions seem most natural also in interaction between humans. On the Pepper robot (120 cm tall), participants preferred that the other hand of the robot remained still (reaction A). For the Nao (60 cm tall), the reaction with the contralateral hand moving simultaneously forward was preferred (reaction B)\footnote{This reaction was identified as one that is taught in courses of self-defense.}. However, the participants' assessments were not clear-cut and the effect of preferring A or B (touched hand moved back) over C or D (touched hand moved front) was only marginally significant. This may have to do with to what extent participants anthropomorphized the robot. When asked to comment on the scenario and their choice of the preferred reaction, some said ``this is what I would do''. This perspective was probably not shared by all participants though---in other words, the most natural reaction one expects from a robot may be different. 

The experiments were conducted on two humanoid robots that differed in several aspects. We speculate that a key factor possibly explaining some of the differences in participants' ratings was the robot size. The Pepper, which is twice as tall as the Nao, may be perceived as potentially more threatening and movements toward a person may not be perceived well---this may explain the preferrence for reaction A in the Pepper. Furthermore, participants with higher scores on the anxiety subscale of the TIPI questionnaire preferred the more passive robot reactions (A,B) (effect not significant). Interestingly, a significant correlation in the opposite direction---more ``anxious, easily upset'' participants preferring reactions C or D---was found for the Nao robot. We speculate that people did not fear the Nao robot; instead more anxious participants were more eager to see a friendly, engaging response of the robot with one or both arms moving forward. 

The robot reactions we designed additionally consisted in leaning back after the unexpected touch, followed by looking at the participant (and blinking in case of Pepper). These were identical in all conditions. In a questionnaire, we asked the participants whether they noticed the leaning back and whether they liked it---the majority of participants responded positively to both questions. Looking at the participant after the contact was not specifically rated; few participants appreciated it in unstructured comments on the experiment. Looking at the touched body part instead of at the face of the participant constitutes an alternative that could be tested in the future. 

We also studied where on the hand and how people contacted the robot. Similarly for both robots, participants typically used the fingers of their left hand to touch the outer or frontal part of the robot hand. Most participants used a brief touch/contact; only few participants used other touch types like pat, grasp, or scratch. 

In summary, this study provides a new contribution to the design of robot responses to touch. In other works on social touch in HRI, experiments end their evaluation with the moment of touch and the reaction to touch and its appropriateness is not assessed (e.g., \cite{alenljung2018,burns2022endowing}). Participants are asked about their perception of the robot (e.g., \cite{arnold2018}) or pleasantness or appropriateness of the situation involving touch in general (e.g., \cite{chen2011touched,cramer2009,willemse2016observing,zheng2020}). Instead, we specifically asked them to rate the robot reactions. Furthermore, unlike in studies
 in which participants are asked to touch a robot in a specific fashion \cite{alenljung2018,burns2022endowing}, we studied the natural way in which participants touch a robot to get its attention. 
 
The implications of this work are the following. First, it seems that high-level behavioral patterns such as ``I move a hand that was unexpectedly touched back'' can be carried over from human-human to human-robot interaction. However, the size of the robot seems to be an important factor, shaping how participants rate the robot reactions. Furthermore, there may be interaction with the personality traits of the interlocutor and hence, if possible, the reactions may be personalized. Second, the analysis of touch location and type revelaed a preference for brief contact with the outer or inner part of the robot hand, which should be taken into account in robot design---placement and type of contact sensors. Finally, our scenario involved deliberate contact. It remains to be tested whether the reactions we designed would be also be positively rated in incidental contact scenarios like when a robot is traversing a crowded place \cite{garcia2019}.

\bibliographystyle{splncs04}
\bibliography{RobotStartle}

\end{document}